\documentclass[10pt,twocolumn,letterpaper]{article}

\usepackage{cvpr}
\usepackage{times}
\usepackage{epsfig}
\usepackage{graphicx}
\usepackage{amsmath}
\usepackage{amssymb}

\usepackage[pagebackref=true,breaklinks=true,letterpaper=true,colorlinks,bookmarks=false]{hyperref}

 \cvprfinalcopy 

\def\cvprPaperID{11} 

\ifcvprfinal\fi
\begin{document}

\title{Improvements and Experiments of a Compact Statistical Background Model}

\author{Dong Liang, Shun'ichi Kaneko\\
Graduate School of Information Science and Technology, Hokkaido University\\
{\tt\small \{liang, kaneko\}@ssc.ssi.ist.hokudai.ac.jp}
}

\maketitle

\begin{abstract}
Change detection plays an important role in most video-based applications. The first stage is to build appropriate background model, which is now becoming increasingly complex as more sophisticated statistical approaches are introduced to cover challenging situations and provide reliable detection. This paper reports a simple and intuitive statistical model based on deeper learning spatial correlation among pixels: For each observed pixel, we select a group of supporting pixels with high correlation, and then use a single Gaussian to model the intensity deviations between the observed pixel and the supporting ones. In addition, a multi-channel model updating is integrated on-line and a temporal intensity constraint for each pixel is defined. Although this method is mainly designed for coping with sudden illumination changes, experimental results using all the video sequences provided on changedetection.net validate it is comparable with other recent methods under various situations.
\end{abstract}

\section{Introduction}
Object detection in real world scenes involves dealing with illumination changes and moving backgrounds.
Recent studies focus on introducing sophisticated statistical models to describe challenging scenes for background subtraction, from the earliest single Gaussian model \cite{Wren} at each pixel, to the Mixture of Gaussian (MoG) \cite{bouwmans2008background}, and the non-parametric technologies \cite{vibe, Elgammal, kim2005real, wang2006background}. On the other hand, background models or local features which can represent spatial information have shown great potential \cite{heikkila2006texture,learned2012background, liao2010modeling, sheikh2005bayesian}, even if considering spatial information among pixels may increase the time and space complexity of an algorithm. 
Since background modelling is typically the first step integrated in a specific intelligent surveillance system, following much higher-level detection-based tracking, object recognition or identification tasks in a field of view, such pipeline framework generally require a relative succinct background model to reduce the occupancy of computing resources especially when implementing real-time processing, implanting a embedded system, or using multi-camera sensing.

This paper reports the modification and its experimental performance of a single-Gaussian-based model: Co-occurrence Probability-based Pixel Pairs (CP3) \cite{liang, liang1}. CP3 is originally proposed  for off-line object detection under sudden illumination changes, and it is proved to be robust in sudden illumination changes, weak illumination, and regular dynamic background. In this paper, we expand it to be capable of on-line training and detection (Section 3). We mainly modified three parts: multichannel colour model, to better distinguish objects from camouflage effects (photometric similarity of object and background) than just using grayscale information; A model parameters on-line updating is proposed; a temporal intensity range constraint of a pixel is integrated. In section 4, we describe the experiment procedure, parameter setting, and experimental results using changedetection.net dataset \cite{cdw}, which validate its comprehensive performance is comparable with other methods under 11 different situations. 
\section{CP3 Background model}
\begin{figure*}[t]
\begin{center}

   \includegraphics[width=\linewidth]{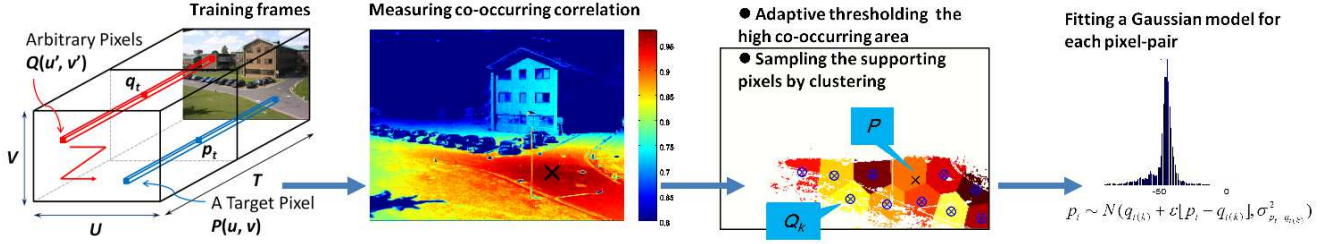}
\end{center}
   \caption{Schematic diagram of CP3 background model using PATS2001-dataset3-cam1 as a demonstration.}
\label{f1}
\end{figure*}
Fig.\ref{f1} is a schematic diagram of CP3 background model.
Given a time series, the intensities of a pixel $P$ have different variation from other arbitrary pixel $Q$ as time goes by. While, it is also natural that $P$ would have some co-occurrence character with some other pixels, for example, has a stable  difference value with its neighbouring pixels. When such kind of co-occurring relation is relative steady, we can fit the deviation of the pixel pair as a single Gaussian. To reduce the risk of individual error and perform robust detection, it is necessary to maintain sufficient number of $Q$ with scattered locations as supporting pixels, denoted as $\{Q^P_k\}_{k=1,2,...,K}$, which provides a group of estimation for $P$ via an unique Gaussian of each pixel pair, where only two parameters, the mean value $\mathcal{E}[p_t-q_{t(k)}]$ and the standard deviation ${\sigma_{p_t-q_{t(k)}}}$, are recorded for the following detection procedure. Once the true intensity of $P$ is far from the background model, $P$ would be regarded as an abnormal-status/foreground-element.

Computing the co-occurring relation of an arbitrary pixel pair uses a covariance-based correlation matrix. Each row and column of the symmetric matrix is an array of correlation coefficient $\gamma_{(P,~Q)}$ for each $P(u,v)$. Then $Q_n$ corresponding to the highest $N$ components in the array $\gamma_{(P,~Q(u',v'))}$ can be selected as the candidates of preferred supporting pixels, namely 
$\{Q_n\}=\{Q(u',v')|\gamma_{(P,~Q)}>\check{\gamma}\}, \ {n=1,2,...,N}$.
where $\check{\gamma}$ is the adaptive lower limit, which can be determined by a computable ratio of the signal variance $\sigma^2_{p_t}$ and the random noise variance $\sigma^2_n$ \cite{liang1}.

The location of supporting pixel depends on the character of its target pixel. For a target pixel on a static background, it finds supporting pixels on the static area (this is the most comment case, the supporting pixels randomly locate on the scene); For a pixel under illumination changes, high co-occurring supporting pixels typically distribute around it, following the illumination motion, and also relate to the geometrical characteristic (position, orientation, shape and relative distance); For a pixel passed by regular motion, it can have the supporting pixels which have simultaneous motion with it, and they locate along the vertical direction of motion.

The next step of CP3 is to sample $K$ number of $\{Q^P_k\}$ with scattered locations from $\{Q_n\}$ for each $P$, then build the underlying Gaussian model. Note that, similar mean value is not a necessary condition for co-occurring pixel pairs, even a pixel-pair which shows a clear intensity difference, is possible, and also can be a qualified co-occurring pixel pair (e.g. the pixel on the road with a low intensity value and the one on the grass with a high intensity value). In the detection stage, using a probability function to measure how much proportion of the pixel pairs be out of the underlying background model.

In summary, this method takes into account of consistency of a pixel pairs, which can bear illumination changes and regular dynamic background. The consistency of a pixel pair provides a simpler probability distribution than a pixel's probability distribution, which is easier to estimate model's parameters, and also easy to set the detection parameters to get reliable results. And, this method also takes into account of the supporting pixels' scattered distribution, which involve spatial sampling (in \cite{liang}, distance-based spatial clustering are used to select scattered supporting pixels). This is an significant improvement to overcome a weakness of a previous study \cite{gap}. In \cite{gap}, no mechanism is used to select scattered supporting pixels, often resulting a more dense supporting pixels cluster. Once such cluster covered by unexpected (untrained) motion at the same frame, large area of false positive detections would appear. CP3 avoids most of such case because the scattered distribution of supporting pixels limits the diffusion of a local ill-condition. In addition, this method also takes into account of randomization of the supporting pixel's location. CP3 uses a random sampled mechanism to select supporting pixels (in \cite{liang} the K-means initialization is random, resulting random samples of supporting pixels at different time), which means even for two neighbouring and homogeneous pixels, the locations of their supporting pixels can be very different. Such kind of scattered and random supporting pixels effectively reduce the risk of false positive, at least most of them are decomposed into sparse noise, which is facile to be removed by post-processing. The opinion of randomization is reasonable for background subtraction, because no  background model guarantee an exactly description of a current frame according to the processing of historic frames or spatial relations. The opinion of randomization is also used for background modelling by \cite{vibe,PBAS} and proved to be effective. 

However, since CP3 is originally designed for off-line object detection, the consistency assumption of the pixel pairs would be broken when using a unlearned video for a longer time. Hereafter, we modified this method to adapt to more general on-line training and detection. We do not expect better performance than a off-line way because on-line training often have deficient priori knowledge. We expect the modification can preserve the merits of CP3 at background initialization, and then introduce new mechanism to better adapt to new coming frames.

\section{Improvements}

To better distinguish objects from camouflage effects, we firstly introduce colour vector instead of grayscale observations. We define $\bf{p_t}$ and $\bf{q_{t(k)}}$ to be the colour vector on RGB colour space, and $\bf{\Delta}_{t(k)}$ to be the mean of $(\bf{p_t}- \bf{q_{t(k)}})$, and $\bf{\Sigma}_{t(k)}$ to be the corresponding covariance matrix. In each recent frame, we update each pixel-pair's statistics recursively. For mean value,
\begin{equation} 
\bf{\Delta}_{t(k)}=\alpha(\bf{p_t}-\bf{q_{t(k)}})+(1-\alpha)\bf{\Delta}_{(t-1)(k)}.
\label{e1}
\end{equation} 
For covariance matrix,
\begin{multline}
$$\bf{\Sigma}_{t(k)}=\alpha(\bf{p_t}-\bf{q_{t(k)}}-\bf{\Delta}_{t(k)})(\bf{p_t}-\bf{q_{t(k)}}-\bf{\Delta}_{t(k)})^T\\
+(1-\alpha)\bf{\Sigma}_{(t-1)(k)}$$.
\label{e2}
\end{multline}
We use a blind updating policy that just add the new sample to the model regardless of it is classified as a background or foreground. Generally speaking, in order to adapt quick background change, we should use a relative large updating rate $\alpha$. But blind updating with large updating rate would produce object's ``tail'' . In the following experiments, we set $\alpha$ to be a small value (Tab.~\ref{t1}), because in CP3 method, the status of a target pixel $P$ is represented jointly by a group of supporting pixels, such representation can deal well with quick illumination change. The updating formulas Eq.~(\ref{e1}) and  Eq.~(\ref{e2}) are to adjust each pixel-pair's model parameter, rather than deal with quick change. Another objective to use blind updating is that it avoids a risk that an object is deadlocked as a foreground for a long time.

Besides model updating, another signification modification is introducing the minimum $\bf{\hat{p}_t}$ and maximum $\bf{\check{p}_t}$ colour values to represent a dynamic range of a background pixel. This step is similar to  \cite{kim2005real}.  The motivation of introducing such mechanism is to overcome a weakness of CP3: the Gaussian model is to model the relative relation of a pixel pair (pixel pair's deviation), but ignore a pixel's intensity range. When processing a short-distance surveillance video, the object is possible to be quite large and cover target pixel and most of the supporting pixels, once the object's texture is similar to the background's (for example, both are smooth with poor texture), it would result large number of false negative, because the passed object would not break the pixel pair's deviation relation, even if its intensity has obvious change. $\bf{\hat{p}_t}$ and $\bf{\check{p}_t}$ are also updated with other two parameters together, the details of their updating are same as \cite{kim2005real}. We combine the upper and lower thresholds with the Gaussian constrain together, and use logic AND operation to integrate them.   
\begin{table}[b]
\begin{center}
\begin{tabular}{lc}

\hline
Number of supporting pixels $K$ &	20\\	
Probability function threshold $pf$ &	0.35\\		
Gaussian model threshold $C$ &	3.0\\
Updating rate $\alpha$ &	0.01\\
\hline
\end{tabular}
\end{center}
\caption{Parameter setting.}
\label{t2}
\end{table}

\section{Experiments}
While using CP3 to off-line process a given dataset, an ideal way is to use a set of training samples covering as much as the background changes of the entire video. We can expect that the detection performance using small number of initial training samples will decrease the robustness of the model. However, without losing generality in real world applications and in order to fairly compare with other background subtraction methods on CDW-2014, we just use the first 100 frames of every dataset for model initialization, and then use the mentioned on-line way to detect object and update background model.  All used parameters are listed in Tab.~\ref{t2}, and a detailed discussion of parameters can be found in \cite{liang1}. Note that, we use different $pf$ and $C$ rather than the values in \cite{liang1} that we hope to reduce the false positive rate globally. The changes of $pf$ and $C$ provide a higher tolerance to different dynamic background but would reduce model's sensitivity. But in the modification we have a minimum and maximum to restrain the dynamic range of a pixel, which helps to maintain sensitive detections. In addition, we have no parameter to deal with cast shadows, although it is possible to simple integrate other studies to suppress cast shadows \cite{cucchiara2003detecting, Elgammal, prati2003detecting}, which use dual (or more) thresholds to restrain the range on an appropriate colour space. However, because the appearance of cast shadow in reality varies depend on different illumination situation, imaging sensor and scenario, those threshold-based methods are not so robust under different cases, for example, in dark area of the image, colour analysis typically work poorly.  More complicate methods involve the combination of colour, geometrical and temporal characteristics \cite{sanin2012shadow} which is out of our scope.

We test all 11 categories in CDW-2014, Tab.~\ref{t1} gives the category-wise quantitative analysis using the 7 metrics provided by the workshop utilities based on True
Positive (TP), True Negative (TN), False Positive (FP), and False Negative (FN). For overall results, our method stands at the middle position compared with other 6 methods listed on CDW-2014 available at the beginning of April, the average ranking is 3.43/7. More detailed about specific metrics, the average Recall= TP/(TP+FN)=0.7225 and False Negative Rate (FNR) =FN/(TP+FN)=0.2775, stands at the second position. Recall and FNR are just the inverse of each other to present the completeness of the foreground after detection, which means our method can preserve well completeness of the foreground blob. We also pay attention to  Percentage of Wrong Classifications (PWC) =100*(FN+FP)/(TP+FN+FP+TN) and F-Measure =2*Precision*Recall)/(Precision+Recall), and PWC should be as small as possible and F-Measure just the reverse. PWC and F-Measure of our method is better than most of other methods (PWC=3.4318\% and F-Measure=0.5805), and stands at the second position. We will do further comparison and deeper discussion after more methods are submitted to CDW-2014.
For computation time, a Matlab code for on-line updating and detection of the modified method can reach around 20 fps on an Intel i7 PC.

\begin{table*}
\begin{center}
\begin{tabular}{|l|c|c|c|c|c|c|c|}

\hline
Categories &	Recall	&	Specificity	&	FPR		&	FNR		&	PWC		&	Precision	&	F-Measure\\
\hline\hline		
\textit{6 categories in CDW-2012}&&&&&&&\\
baseline &0.8500&	0.9972&	0.0027&	0.1499&0.7724&	0.9251&	0.8856\\

dynamicBackground&0.6851&	0.9876&	0.0124&	0.3149&	1.5025&	0.5159&	0.5353\\
cameraJitter&0.6628&	0.9518&	0.0481&	0.3371&	5.9332&	0.4561&	0.5207\\
intermittentObjectMotion&0.7825&	0.8745&	0.1254&	0.2174&	11.5284&	0.5630&	0.6176\\
shadow&0.7839&	0.9832&	0.0167&	0.2160&	2.5175&	0.6539&	0.7036\\
thermal&0.8229&	0.9893&	0.0106&	0.1770&	1.6973&	0.7663&	0.7917\\
\hline\hline
\textit{Other 5 categories in CDW-2014}&&&&&&&\\
PTZ&0.5694&	0.9744&	0.0255&	0.4306&	2.9299&	0.2173&	0.2794\\
badWeather& 0.8350&	0.9954&	0.0045&	0.1649&	0.7411&	0.7423&	0.7766\\
lowFramerate&0.6627&	0.9965&	0.0035&	0.3373&	1.3753&	0.6699&	0.5549\\
nightVideos&0.6308&	0.9470&	0.0530&	0.3692&	6.0131&	0.2775&	0.3483\\
turbulence&0.6885&	0.9948&	0.0051&	0.3114&	0.6646&	0.3950&	0.4724\\
\hline\hline
Overall&0.7225&	0.9705&	0.0295	&0.2775	&3.4318&0.5559	&0.5805	\\
Ranking in \textit{CDW-2014} in Apirl&2/7&5/7&5/7&2/7&2/7&6/7&2/7 \\
\hline
\end{tabular}
\end{center}
\caption{Experiment results of CP3 using changedetection.net 2014 dataset.}
\label{t1}
\end{table*}


\section{Conclusions}

This paper reports the modification of CP3 for on-line change detection and its performance using video sequences provided on CDW-2014. The experimental results under all 11 different categories validate its comprehensive performance is comparable with other change detection methods.

{\small
\bibliographystyle{ieee}
\bibliography{arxiv}
}

\end{document}